%% file: 0main.tex
\documentclass[runningheads]{llncs}

\usepackage{eccv}

\usepackage{eccvabbrv}

\usepackage{graphicx}
\usepackage{booktabs}
\usepackage{multirow}
\usepackage{algorithm}
\usepackage[noend]{algpseudocode}

\usepackage[accsupp]{axessibility}  

\usepackage[pagebackref,breaklinks,colorlinks,citecolor=eccvblue]{hyperref}

\usepackage{orcidlink}

\begin{document}

\title{Low-Frequency Black-Box Backdoor Attack via Evolutionary Algorithm} 

\titlerunning{Abbreviated paper title}

\author{Yanqi Qiao\inst{1} \and
Dazhuang Liu\inst{1} \and
Rui Wang\inst{1} \and Kaitai Liang\inst{1}}

\authorrunning{F.~Author et al.}

\institute{Delft University of Technology, Delft, Netherlands\\
\email{\{y.qiao,d.liu-8,r.wang-8,kaitai.liang\}@tudelft.nl}}

\maketitle
\input{1abstract.tex}
\input{2introduction}

\input{3related_work}
\input{4method}

\input{5experiments}

\input{6conclusion}

\clearpage

%
%
\bibliographystyle{splncs04}
\bibliography{7ref}
\input{8appendix}

\end{document}

%% file: 1abstract.tex
\begin{abstract}

Convolutional Neural Networks (CNNs) that have excelled in diverse computer vision tasks are vulnerable to backdoor attacks, enabling attacker-controlled predictions via specific patterns. 
Restricted to spatial domains, recent research exploits perceptual traits by embedding patterns in the frequency domain, yielding pixel-level indistinguishable perturbations. 
In black-box settings, restricted access to training and model necessitates advanced trigger designs. 
Current frequency-based attacks manipulate magnitude spectra, introducing discrepancies between clean and poisoned data, though vulnerable to common image processing operations like compression and filtering.

In this paper, we propose a robust \underline{l}ow-\underline{f}requency practical \underline{b}ackdoor \underline{a}ttack (\textbf{LFBA}) in black-box setup that minimally perturbs low-frequency components of frequency spectrum and maintains the perceptual similarity in spatial space simultaneously.
Our methodology capitalizes on the insight that optimal triggers can be located in low-frequency regions to maximize attack effectiveness, robustness against image transformation defenses, and stealthiness in \emph{dual} space. 
We utilize simulated annealing (SA), a form of evolutionary algorithm, to optimize the properties of frequency trigger including the number of manipulated frequency bands and the perturbation of each frequency component, without relying on prior knowledge of the victim classifier. 
Extensive experiments on real-world datasets confirm the effectiveness and robustness of LFBA against image processing operations and state-of-the-art backdoor defenses. 
Furthermore, LFBA exhibits inherent stealthiness in both spatial and frequency spaces, making it resistant to frequency inspection.

\keywords{backdoor attack, black-box, frequency domain, simulated annealing, robustness}
\end{abstract}

%% file: 2introduction.tex
\section{Introduction}

CNNs are vulnerable to backdoor attacks \cite{sig,Blended,badnets,Trojaning,refool,input-aware} that can mislead the model to make attack-chosen predictions with triggers in the use phase while behaving normally on clean images, causing severe consequences in high-stakes applications such as autonomous driving \cite{autonomous} and biometric authentication \cite{Biometrics_authentication}. 

\begin{figure}[t]
    \centering\scalebox{0.9}{
    \includegraphics[width=1\textwidth]{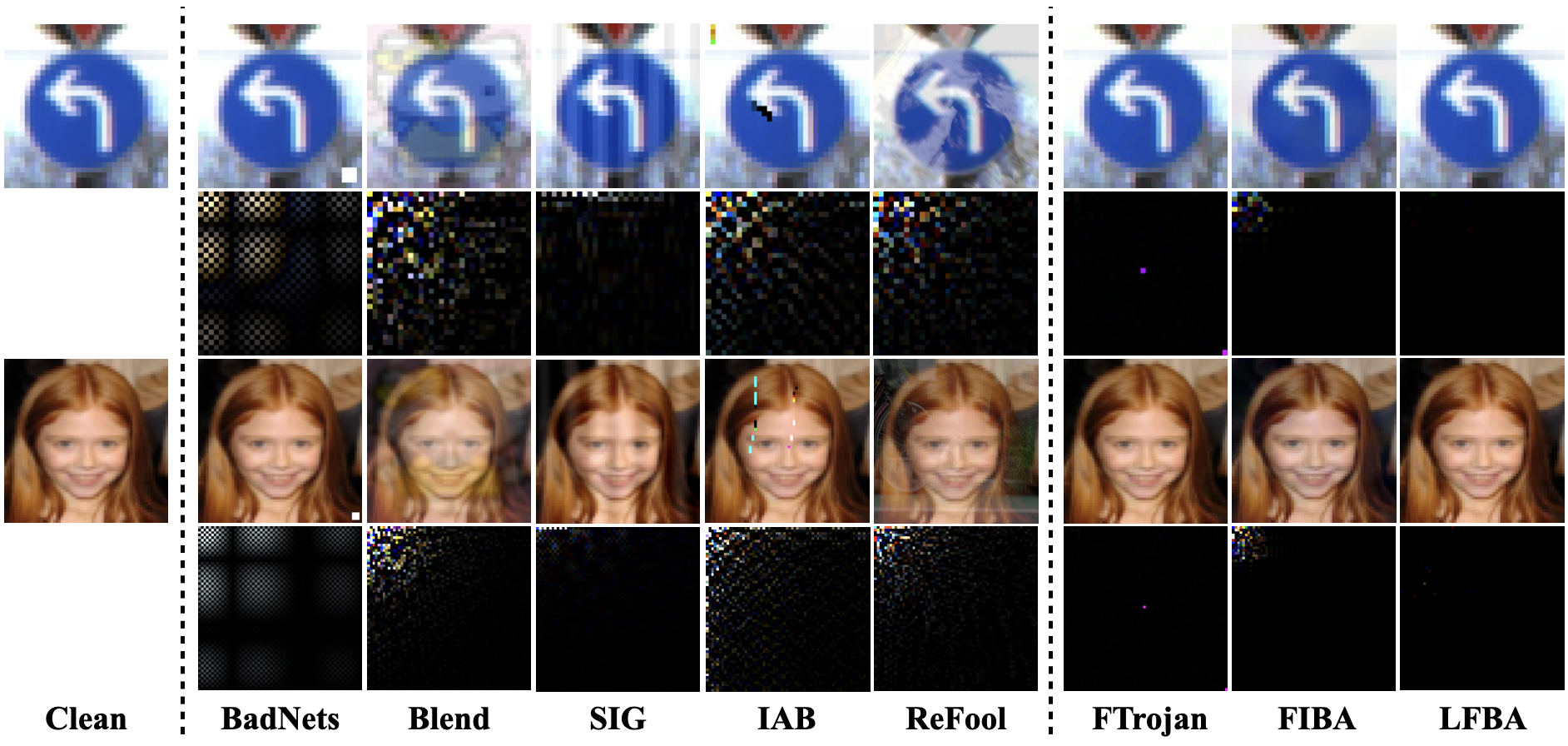}}
    
    \caption{Comparison of poisoned images with their corresponding frequency disparities (amplified by $5\times$) to clean images of existing attacks.
    \textbf{Left}: clean images;
    \textbf{mid}: poisoned images from spatial domain based attacks including BadNets \cite{badnets}, Blend \cite{Blended}, SIG \cite{sig}, IAB \cite{input-aware} and ReFool \cite{refool}; \textbf{right}: poisoned images from frequency domain based attacks including FTrojan \cite{ft}, FIBA \cite{fiba} and our LFBA attack. 
    Although state-of-the-art frequency triggers achieve superior perceptual similarity than spatial triggers, they introduce anomaly frequency artifacts.}
    \label{poison_img}
    \vspace{-0.2in}
\end{figure}

Prior backdoor attacks possess the capability to inject imperceptible triggers into spatial domain \cite{lira,issba,wanet,wb,IBA,defeat,dfst}.
Inserting triggers in spatial space can harm the semantics of infected image pixels (see \Cref{poison_img}).  
Recent works have concluded that backdoor attacks can inject trigger patterns into frequency space \cite{ft,fiba,cyo,stealthy_freq}.
For example, FTrojan \cite{ft} manipulates mid- and high-frequency spectrum of input images with a pre-defined perturbation within fixed frequency band.
Manually crafting frequency components, especially in high-frequency regions, could harm the robustness of trigger and thus trigger effectiveness can be eliminated by image processing operations such as lowpass filters.
Moreover, both spatial and frequency triggers introduce distinguishable artifacts when transformed to frequency space (see \Cref{poison_img} and \Cref{freq_inspection}).

Inspired by \cite{color}, an ideal and practical backdoor attack should achieve four objectives, namely, \emph{functionality preservation}, \emph{effectiveness}, \emph{dual-space stealthiness}, and \emph{robustness}.
Functionality preservation ensures high test accuracy on clean data. 
Effectiveness is demonstrated by the ability to misclassify poisoned data to the target label with a high probability. 
Dual-space stealthiness implies that poisoned images exhibit visual and frequency similarity to clean ones. 
Robustness is demonstrated by its effectiveness against image transformations and resistance to backdoor defenses.
Although successfully achieving the goals at pixel level, \cite{color} does not consider the stealthiness in the frequency perspective.
This work explores a new perspective of attacks in the frequency domain.

Typically, low-frequency components of an image contain semantic information, while the high-frequency components capture finer details and noise.
According to prior works such as \cite{watermark,lfap}, inserting triggers in low-frequency region offers concrete influences: 
(1) low-frequency components have a perceptual capacity that allows trigger insertion without perceptual degradation;
(2) low-frequency components exhibit greater resilience in lossy compression operations such as JPEG, whereas high-frequency components are more pronounced to data loss;
and (3) trigger inserted in low-frequency region is harder to be removed by low-pass filtering compared to the high-frequency region.

Building on the above insights, we develop LFBA, a robust and practical low-frequency backdoor attack with imperceptible triggers in dual space.
The key insight of LFBA is to find the optimal trigger that can achieve both dual-space stealthiness and attack robustness when the model and defense strategy are unknown.
This design marks the first exploration into the robustness from a frequency perspective, which locates mininal perturbations in low-frequency region against image transformations while naturally guaranteeing perceptual similarity in the pixel domain.
Finding such a frequency trigger in a black-box scenario is not trivial. 
Due to the absence of the victim model and training process, one may handcraft the frequency trigger, but it could lead to improper frequency properties of the trigger.
For example, a large perturbation can disrupt invisibility of poisoned images, while a small perturbation could hinder the model's ability to learn the trigger feature, resulting in low attack effectiveness. 
Furthermore, selecting an improper frequency band for trigger insertion can compromise the attack robustness against image processing operations (see \Cref{preprocess_defense}). 

To address the challenges, we leverage simulated annealing (SA), a gradient-free optimziation algorithm, to effectively search for the optimal trigger in frequency domain. 
Specifically, we first convert a clean image to frequency domain with discrete cosine transform (DCT).
We iteratively optimize the perturbations and frequency bands of trigger with SA, in order to maximize the attack effectiveness as the primary goal and a penalty term concerning the dual-space stealthiness.
Due to lacking knowledge about the target model in the black-box setting, we approximate the quality of trigger in terms of backdoor loss in the optimization process with a semi-trained evaluation model.
Finally, we use inverse DCT (IDCT) to produce the poisoned image.
Since the imperceptible perturbations is posed in frequency space, LFBA preserves the invisibility in the spatial domain and is robust to any eradication in frequency domain.
The main contributions of this work are as follows:

\noindent $\bullet$ We explore both stealthiness and robustness of trigger from frequency perspective and design an imperceptible and robust black-box backdoor attack in dual space. 
\\
$\bullet$ We propose a constrained optimization problem to find optimal frequency triggers.
Then we utilize SA to solve the objective without relying on gradient information.
\\
$\bullet$ Extensive experiments empirically demonstrate that the proposed attack provides state-of-the-art effectiveness and robustness against existing spatial and frequency defenses as well as image transformation defenses.

%% file: 3related_work.tex
\section{Related Work}

\noindent\textbf{Backdoor Attacks.} 
\cite{badnets} introduces the first backdoor attack against deep learning models. 
It employs a patch-based pattern as trigger, injecting it into a small fraction of clean data, which causes the victim model to misclassify those poisoned images to the target label.
After that, various attacks, aiming at improving stealthiness and robustness through the design of triggers and training process, have been proposed in the literature.
\\
(1) \emph{Spatial domain-based attacks}: To enhance the invisibility of triggers for bypassing human inspection, some works \cite{sig,refool,lira,issba,wanet,color} focus on imperceptible backdoor attacks in spatial domain. 
For example, \cite{sig} uses sinusoidal signals as triggers which results in slightly varying backgrounds;  
\cite{refool} utilizes natural reflection as trigger into the victim model, while  
\cite{issba} leverages DNN-based image steganography technique to hide an attacker-specified string into clean images as sample-specific triggers.
Later, several works \cite{wb,IBA,defeat,dfst} reveal the importance of stealthiness in latent feature space.
\cite{wb} learns a trigger generator to constrain the similarity of hidden features between poisoned and clean data via Wasserstein regularization. 
To improve the stealthiness of triggers, \cite{defeat} adaptively learns the generator by constraining the latent layers, which makes triggers more invisible in both input and latent feature space. 
While spatial attacks offer inherent stealthiness, they often overlook robustness against common image processing operations utilized during data preprocessing. Consequently, their effectiveness is significantly compromised by such operations.
Moreover, most attacks require a strong attack assumption that the adversary possesses full control over the training process and has knowledge of the victim model. 
More importantly, many spatial backdoor attacks exhibit severe high-frequency artifacts that can be easily detected in the frequency domain (see \Cref{poison_img} and \Cref{freq_inspection}).
\\
(2) \emph{Frequency domain-based attacks}:
Recent works \cite{cyo,ft,fiba,rethink,stealthy_freq} explore another attack surface, namely, frequency domain, naturally guaranteeing invisibility due to frequency properties. 
\cite{ft} handcrafts two single frequency bands with fixed perturbations as trigger, and \cite{fiba} injects low-frequency information of a trigger image by linearly combining spectral amplitude of poisoned and clean images. 
These works introduce distinguishable frequency artifacts and can be detected via frequency inspection.
Moreover, they focus on natural stealthiness and do not consider robustness against image transformation operations. 

Different from the above works, we propose a black-box frequency backdoor attack that firstly achieves imperceptibility in dual spaces and robust against image processing defenses.

\noindent\textbf{Backdoor Defenses.}
Defensive \cite{fine_pruning,nc,deepinspect,generative_distribution_modeling,bridging,nad,Rethinking} and detective \cite{strip,activation_clustering,spectral,ulp,rethink} mechanisms are commonly used for backdoor defenses. 
Defensive methods focus on mitigating the effectiveness of potential backdoor attacks.
For example, fine-pruning \cite{fine_pruning} prunes the dormant neurons in the last convolution layer based on clean inputs' activation values.
Neural Cleanse \cite{nc} reconstructs potential triggers for each target label via reverse engineering and renders the backdoor ineffective by retrain patches strategy.
Neural Attention Distillation \cite{nad} uses a ``teacher" model to guide the finetuning of the backdoored ``student" network to erase backdoor triggers. 
Representative detective methods include STRIP \cite{strip} which perturbs or superimposes clean inputs to identify the potential backdoors during inference time, spectral signature \cite{spectral} using latent feature representations to detect outliers and \cite{rethink} leveraging supervised learning to differentiate between clean and poisoned data in frequency space. 
Besides, image processing-based methods \cite{Rethinking,ft,deepsweep} have been studied, which remove backdoors by image processing transformations. 
In this work, we showcase that the proposed attack can evade representative defenses including frequency inspection, image processing operations and mainstream backdoor defenses.

\noindent\textbf{Threat Model.}
We consider rather realistic black-box scenario as in prior works \cite{dynamic,color,ft} where the adversary, i.e. a malicious data provider, can only inject a limited number of poisoned samples into clean training set for public use.
The attacker should not have control over the training process or have knowledge of the victim model. 
This is a more practical and challenging attack scenario than white-box attacks \cite{dfst,wb,lira,marksman,defeat}.

%% file: 4method.tex
\section{Proposed Method}

\subsection{Preliminaries}
We consider backdoor attacks on image classification. 
Let $f_\theta: \mathcal{I}\rightarrow\mathbb{R}^K$ be an image classifier parameterized with $\theta$ that maps an input image $\mathcal{I}\subseteq [0,1]^{H\times W\times C}$ to an output class, where $K$ is the number of classes, $H$, $W$ and $C$ are the height, width and channels of an input image.
The parameters $\theta$ of the classifier are learned using a training dataset $D_c = \{(x_i,y_i)| x_i\in \mathcal{I}, y_i\in \mathbb{R}^K\}^N_{i=1}$.

In a standard backdoor attack, the attacker crafts a subset of $D_c$ with ratio $\rho$ to produce the poisoned dataset $D_p = \{(x'_i,y'_i)| x'_i\in \mathcal{I}, y'_i\in \mathbb{R}^K\}^{N\times\rho}_{i=1}$ by the trigger function $\mathcal{T}$ and target label function $\eta$. 
Given a clean image $x$ from the clean subset and its true class $y$, the commonly used trigger function $\mathcal{T}$ and target label funtion $\eta$ in the spatial space are defined with a mask $m\in [0,1]$ and a trigger pattern $t$ as follows:
\begin{equation}
    x'=\mathcal{T}(x,m,t)=x\cdot (1-m)+t\cdot m, \quad y'=\eta(y)=y_t, 
\end{equation}
where $y_t$ is the target class.
Under empirical risk minimization (ERM), a typical attack aims to inject backdoors into the classifier $f$ by learning $\theta$ with both clean dataset $D_c$ and poisoned dataset $D_p$ so that the classifier misclassifies the poisoned data into the target class while behaving normally on clean data as follows:
\begin{equation}
    \underset{\theta}{\operatorname{min}}  \sum_{(x,y)\in {D_c}} \mathcal{L}(f_{\theta}(x),y) + \sum_{(x',y')\in {D_p}} \mathcal{L}(f_{\theta}(x'),y'),
\end{equation}
where $\mathcal{L}$ represents the cross-entropy loss.

\subsection{Frequency Backdoor Attack}
We redesign the trigger function $\mathcal{T}$ in the frequency space to better search the frequency trigger that can achieve dual-space stealthiness.
Given a clean sample $(x,y)$ in $D_c$, we first transform it to the frequency domain via DCT function $\mathcal{D(\cdot)}$ and obtain the frequency form $x^f(h^f,w^f,c)$ of input $x(h,w,c)$ as:
\scriptsize\begin{equation}
\mathcal{D}(x(h,w,c)) = V(h^f)V(w^f)\sum_{h=0}^{H-1}\sum_{w=0}^{W-1}\sum_{c=0}^{C-1} x(h,w,c)\cos{\left[\frac{(2h+1)h^f\pi}{2H}\right]}\cos{\left[\frac{(2w+1)w^f\pi}{2W}\right]} 
\end{equation}
\normalsize
for $\forall h^f=0,1,...,H-1$ and $\forall w^f=0,1,...,W-1$, where $H,W,C$ represent the height, width and number of channels of the image. 
For simplicity, we assume $H=W$, therefore $V(0)=\sqrt{\frac{1}{4H}}$ and $V(k)=\sqrt{\frac{1}{2H}}$ for $k>0$.  
Accordingly, $\mathcal{D}^{-1}(\cdot)$ denotes the IDCT as follows:
\scriptsize\begin{equation}
\mathcal{D}^{-1}(x^f(h^f,w^f,c)) = \sum_{h^f=0}^{H-1}\sum_{w^f=0}^{W-1}\sum_{c=0}^{C-1} V(h)V(w)x^f(h^f,w^f,c)\cos{\left[\frac{(2h^f+1)h\pi}{2H}\right]}\cos{\left[\frac{(2w^f+1)w\pi}{2W}\right]} 
\end{equation}
\normalsize
Therefore, our frequency trigger function is defined as:
\begin{equation}
    \mathcal{T}^f(x, \delta, \nu)= \mathcal{D}^{-1}(\mathcal{D}(x)_{\nu} + \delta_{\nu}),  
\label{trigger_function}
\end{equation}
where the trigger $\delta_\nu$ comprises a set of perturbations $\delta=\{\delta_i|i=1,2, \dots, n\}$ in terms of frequency components and its corresponding frequency bands $\nu=\{\nu_i | i=1,2, \dots, n\}$ that indicates the position in frequency spectrum to pose the perturbation $\delta$ on, $n$ represents the number of frequency bands to manipulate. 

Our primary goal is to search the optimal trigger that can achieve high attack effectiveness.
Ideally, the effectiveness of trigger should be assessed by using the victim classifier, whereas the adversary lacks knowledge about it in practical attack scenario.
As an alleviation, we utilize the poisoned set to learn a semi-trained evaluation classifier $f^s$, which may be derived from open-source architectures.
This evaluation classifier can approximate the trigger effectiveness with an acceptable deviation in practice.  
Therefore, the main task of our attack to minimize is defined as follows:

\begin{equation}
O(\delta, \nu)= \sum_{(x,y)\in {D_p}}  \mathcal{L}(f^s_{\theta}(\mathcal{T}^f(x, \delta, \nu), y_t)
\label{loss}
\end{equation}

One may argue that large crafted perturbations in specific frequency bands could also achieve high attack effectiveness and practical natural stealthiness such as \cite{ft,fiba}.
Frequency triggers without careful consideration can bring distinguishable artifacts in the frequency space (see \Cref{poison_img} and \Cref{freq_inspection}).
We hereby use dual-space stealthiness penalties to ensure imperceptible perturbations and invisibility of poisoned images in both frequency and spatial space. 
We define penalty term for dual-space stealthiness as follows\footnote{The distance between clean and poisoned samples remains consistent between frequency and spatial domains. We hereby only constrain frequency disparities.}:
\begin{equation}
    P(\delta,\nu)= \|\mathcal{D}(\mathcal{T}^f(x, \delta, \nu)-x)\|_p
\label{stealthiness}
\end{equation}
where $p=2$ denotes $l_2$-norm distance used to calculate the disparity between clean and poisoned sample in the frequency domain.
Taking into account both the primary goal and the dual-space stealthiness penalty, our attack goal aims to minimize the overall objective function under the constraint w.r.t. frequency perturbation.
The optimization problem is formulated as follows:

\begin{equation}
\begin{aligned}
\min_{\delta,\nu} \quad & O(\delta,\nu)+P(\delta,\nu) & \\
\mbox{s.t.}\quad
&\|{\delta_\nu}^{(i)}\|\leq\epsilon  & \\
\end{aligned}
\label{eq_objectiveFunction}
\end{equation}
where \emph{i}$\in$ $\left[1,\emph{n}\right]$, $n$ is the number of manipulated bands, and $\epsilon$ is the maximal perturbation for each frequency band in LFBA attack.

\subsection{Frequency Trigger Optimization via Simulated Annealing}

We learn trigger with simulated annealing (SA) \cite{van1987simulated}, a commonly used probabilistic based optimization technique without relying on gradient information. 
The annealing process involves heating a material to a high temperature and then gradually cooling it to remove defects and optimize its internal structure. 
To reflect this process, we decrease the temperature \emph{T} from $T_{0}$ to $T_{\emph{f}}$, as denoted in \emph{Step 2} in \Cref{alg_SA}, to control the trigger optimization process. Under each temperature \emph{T}, frequency band and perturbations are changed to minimize the attack objective value calculated in \cref{eq_objectiveFunction}. 

\begin{algorithm}[t]
    \caption{Optimal frequency trigger search via SA}
    \begin{algorithmic}[1]
        \Require{Poisoned dataset $\mathcal{D}_p$, Initial temperature $T_{0}$, Terminal temperature $T_{f}$, 
        Optimization iterations per temperature \emph{iter}, 
        Annealing factor \emph{$\alpha$}, 
        Maximum frequency perturbation \emph{$\epsilon$},  
        Number of retrain epoch $E_{re}$,
        Semi-trained evaluation model $f^s_{\theta}$}
        \Ensure{The optimal frequency perturbations and their bands $\delta^*$, $\nu^*$}
        \State \emph{\boxed{Step\,1: Initialization}}
        \State $\delta_{opt},\nu_{opt} \gets Initialize\_Trigger$
        \State $Obj_{opt}$ = $O(\delta_{opt},\nu_{opt})+P(\delta_{opt},\nu_{opt})$ with $\mathcal{D}_{p}$ on $f^{s}_{\theta}$ 
        \State \emph{T} = \emph{$T_{0}$}
        \State \emph{\boxed{Step\,2: Trigger\ optimization}}
        \While{\emph{T} $\geq$ \emph{$T_{f}$}}
            \For {$i = 1, 2, \dots, iter$}
            \State $\delta \gets $ Rand($\epsilon$), $\nu \gets $ Rand($h_{LF}$, $w_{LF}$) 
            \State Poison $\mathcal{D}_p$ with $\delta, \nu$ using $\mathcal{T}^f$ in \cref{trigger_function}
            \State $f^{s}_{\theta ^{'}}$ $\gets$ Train $f^s_\theta$ on $\mathcal{D}_p$ within $E_{re}$
            \State $Obj \gets O(\delta,\nu)+P(\delta,\nu)$ with $\mathcal{D}_p$ on $f^s_{\theta ^{'}}$
            \If{$Obj > Obj_{opt}$}
                \State $\delta_{opt} \gets \delta, \nu_{opt} \gets \nu$,  $Obj_{opt} \gets Obj$
            \EndIf
            \EndFor
            \State \emph{T} = \emph{T} - $\alpha \times T$
        \EndWhile
        \State \Return $\delta^{*} \gets \delta_{opt},\nu^{*} \gets \nu_{opt}$
    \end{algorithmic}
    \label{alg_SA}
\end{algorithm}

\Cref{alg_SA} describes the workflow of searching our optimal trigger with SA in low-frequency region $(h_{LF},w_{LF})$. 
Particularly, it starts with randomly initializing a trigger in frequency domain. 
After that, SA iteratively improves the trigger stealthiness and effectiveness guided by the objective function in \cref{eq_objectiveFunction} until reaching the termination criteria (i.e., the temperature T drops to the termination temperature $T_{\emph{f}}$). 
Specifically, in each iteration, an offspring $\delta^{'}$ is generated under the constraint in \cref{eq_objectiveFunction} and injected to the clean image through frequency domain. 
Meanwhile, frequency band $\nu$ is also randomly altered. 
Once the new trigger is better than the previous one in terms of the objective value, as calculated in line 4 (for cold start) and 12 (during the iteration) of \cref{alg_SA} , the new trigger will be adopted, otherwise the above process will be repeated.
Finally, the current outperformed optimal trigger among $\delta$ and $\delta_{opt}$ will survive as the new $\delta_{opt}$ and enter the next round. 
Upon termination, the last $\delta_{opt}$ is the desired optimal trigger that is used by \cref{trigger_function} to produce the poisoned dataset.

%% file: 5experiments.tex
\section{Experiments}

\subsection{Experimental Setup}
\noindent\textbf{Datasets and Models.}
Without loss of generality, we evaluate LFBA on five benchmark tasks including handwritten digit recognition on MNIST \cite{mnist}, object classification on CIFAR-10 \cite{cifar10} and Tiny-ImageNet (T-IMNET) \cite{tiny}, traffic sign recognition on GTSRB \cite{gtsrb} and face attribute recognition on CelebA \cite{celeba}.
For CelebA, we follow \cite{dynamic,wanet} to select the top three most balanced attributes including Heavy Makeup, Mouth Slightly Open, and Smiling. 
Then we concatenate them to create an eight-label classification task.
We test LFBA on both small and large-scale datasets with a wide range of image sizes, including both grayscale and RGB images, to verify attack performance and also remain consistency across different types of image datasets. 
Following \cite{bypassing,spectral,activation_clustering,wanet,marksman}, we consider various architectures for the image classifier. 
Specifically, we employ a CNN model \cite{wanet,marksman} for MNIST, PreAct-ResNet18 \cite{resnet} for CIFAR-10 and GTSRB, and ResNet18 \cite{resnet} for T-IMNET and CelebA.
The details of computer vision tasks, datasets and models are described in \Cref{task} (\Cref{appendix:task}).

\noindent\textbf{Evaluation Metrics.}
We evaluate attack effectiveness based on attack success rate (ASR), i.e. the ratio of poisoned samples successfully misclassified to the target label, and test accuracy (ACC) on clean data for functionality-preserving requirement. 
For human inspection, we use PSNR \cite{psnr}, SSIM \cite{ssim} and  LPIPS \cite{lpips} to evaluate spatial invisibility between clean and poisoned data.
LPIPS utilizes deep features of CNNs to identify perceptual similarity, while SSIM and PSNR are calculated based on the statistical pixel-wise similarity.

\noindent\textbf{Implementations.} For the default setting, we train the classifiers by SGD optimizer with learning rate $0.01$ and decayed by a factor of 0.1 after every 50 epochs. 
We set batch size to 64 and total number of epochs to 200. 
We set $\epsilon$ to 0.1 for MNIST, 0.5 for CIFAR-10, GTSRB, and 1.5 for T-IMNET and CelebA.
Following the approach outlined in \cite{lfp}, we select approximately 18.3\% of the frequency spectrum in the top-left region to search for our low-frequency trigger.
We choose $n=3$ as the number of manipulated frequency bands.
For simplicity, we set the poison ratio to only 5\% and target label to 7 for all the datasets\footnote{Our attack is label-independent, i.e., the attacker can easily transfer LFBA attack to any other desired labels by searching the corresponding optimal triggers.}.
Unless explicitly stated otherwise, we adopt this default setting for LFBA in the subsequent sections.  
The implementation of LFBA is based on PyTorch \cite{pytorch} and executed on a workstation with 16-core AMD Ryzen 9 7950X CPU, NVIDIA GeForce RTX 4090 and 64G RAM.
 
\subsection{Attack Performance}

\begin{table}[t]
\centering
\caption{Attack performance via ACC (\%) and ASR (\%) for several attaks.}
\label{attack_performance}
\scalebox{0.9}{
\begin{tabular}{@{}ccccccccccc@{}}
\toprule
      \multirow{2}{*}{Attack} & \multicolumn{2}{c}{MNIST} & \multicolumn{2}{c}{GTSRB} & \multicolumn{2}{c}{CIFAR-10} & \multicolumn{2}{c}{Tiny-ImageNet} & \multicolumn{2}{c}{CelebA} \\ \cmidrule(l){2-3} \cmidrule(l){4-5} \cmidrule(l){6-7} \cmidrule(l){8-9} \cmidrule(l){10-11}
        
  & ACC           & ASR        & ACC           & ASR        & ACC            & ASR          & ACC    & ASR   & ACC    & ASR           \\ \midrule
Clean   & 99.41   & -    & 98.55    & -   & 93.14          & -            & 54.60        & - & 79.20    & -  \\ \midrule
\textsc{BadNets}  & 99.35    & \textbf{99.99}      & 97.91     & 96.67          & 92.05      & 98.24   &  51.90    & 97.82  &  76.54    & 99.35  \\
\textsc{SIG}     & 99.31         & 99.85          & 97.90         & 99.87          & 92.14      & \textbf{99.98}      & 51.98    & 99.49   &  77.90    & 99.85           \\
\textsc{Refool}  & 98.71         & 98.28          & 97.94         & 98.51          & 91.09    & 97.03       & 48.37      & 97.32 &  77.53    & 98.09             \\
\textsc{WaNet}   & 98.59         & 97.09          & 98.19        & 99.83          & 92.31     & 99.94     & 52.85     & 99.16 &  77.99    & 99.33       \\
\textsc{FTrojan} & 99.36         & 99.94          & 96.63         & 99.25          & 92.53  & 99.82   & 53.41       & 99.38  &  76.63    & 99.20            \\
\textsc{FIBA}    & 99.37         & 99.88          & 96.73         & 98.88          & 91.13   & 68.83   &   51.11   & 92.14 & 75.90   & 99.16  \\
\textsc{LFBA}    & \textbf{99.39}   & 99.72   & \textbf{98.42}   & \textbf{99.97}     & \textbf{92.91}    & 99.88     & \textbf{53.64}             & \textbf{99.90} &  \textbf{78.79}    & \textbf{99.91}       \\ \bottomrule
\end{tabular}}
\vspace{-0.2in}
\end{table}

We compare both spatial and frequency attacks including BadNets \cite{badnets}, SIG \cite{sig}, ReFool \cite{refool}, WaNet \cite{wanet},  FTrojan \cite{ft} and FIBA \cite{fiba} as baselines to evaluate the effectiveness.
Since other backdoor attacks \cite{dfst,wb,IBA,lira,defeat} require full control over training process and knowledge of the victim classifiers, we do not consider the above methods as practical baselines. 

\noindent\textbf{Attack effectiveness.}
\emph{We first demonstrate that LFBA achieves high ASR ($\geq$ 99\%) across 5 datasets and 3 models with slight accuracy degradation ($<$0.55\% in average) (see \Cref{attack_performance}).}
The results confirm that our attack outperforms other black-box attacks in most tasks.
That is so because we approximate the trigger effectiveness during the optimization process of SA (see \cref{eq_objectiveFunction}) whereas others do not take into account attack effectiveness when designing their triggers.
Our experimental findings raise urgent concerns for the physical realm: adversaries can compromise any architecture by injecting publicly available images with imperceptible triggers in dual space, even without access to the victim classifier.

\noindent\textbf{Natural stealthiness.}
A dual-space stealthiness penalty is added to the process of searching the optimal LFBA trigger so as to ensure natural stealthiness of poisoned images.
We compare the state-of-the-art invisible attacks in spatial and frequency domains.
For each dataset, we randomly select 500 sample images from test dataset to evaluate trigger stealthiness. 
A higher PSNR/SSIM or a smaller LPIPS value indicates a better stealthiness of an given poisoned sample. 
LFBA achieves more natural stealthiness than current frequency backdoor attacks (see \Cref{natural_stealthiness}) due to its less number of frequency bands and minimal perturbations.
Such minor alterations of LFBA in frequency space can naturally provide invisibility to the potential defender who lacks knowledge of the correspondent clean image. 

Taking \Cref{natural_stealthiness} and \Cref{poison_img} into consideration, we conclude that \emph{the proposed LFBA attack outperforms both spatial and frequency domain-based attacks in terms of natural stealthiness.}

\begin{table}
\vspace{-0.2in}
\centering
\caption{Natural stealthiness (PSNR $\uparrow$, SSIM $\uparrow$, LPIPS $\downarrow$).}
\label{natural_stealthiness}
\scalebox{0.8}{
\begin{tabular}{@{}ccccccccccccc@{}}
\toprule
\multirow{2}{*}{Attacks} & \multicolumn{3}{c}{GTSRB} & \multicolumn{3}{c}{CIFAR-10} & \multicolumn{3}{c}{Tiny-ImageNet} & \multicolumn{3}{c}{CelebA} \\ \cmidrule(l){2-4} \cmidrule(l){5-7} \cmidrule(l){8-10} \cmidrule(l){11-13}
   & PSNR   & SSIM   & LPIPS   & PSNR    & SSIM  & LPIPS    & PSNR    & SSIM      & LPIPS  & PSNR  & SSIM   & LPIPS   \\ \midrule
Clean                   &    Inf    &   1.0000     &    0.0000     &    Inf     &    1.0000     &     0.0000     &    Inf       &     1.0000      &     0.0000  &  Inf  &  1.0000   &  0.0000    \\ \midrule
\textsc{BadNets}                  &    27.18    &     0.9754   &     0.0059    &    36.67     &    0.9763     &     0.0012     &      36.35     &    0.9913       &     0.0006  &  32.50  &  0.9951   &  0.0005    \\
\textsc{SIG}                     &    25.32    &    0.7313    &   0.0766      &    25.26     &     0.8533    &    0.0289      &     25.36      &      0.8504     &     0.0631   &  25.38  &  0.7949   &  0.0359   \\
\textsc{Refool}                  &    20.57    &     0.7418   &    0.3097     &    18.37     &    0.6542     &      0.0697    &    20.42       &      0.8564     &      0.4574 &  23.72  &  0.8359   &  0.2134    \\
\textsc{WaNet}                   &    30.11    &    0.9669    &   0.0584      &    19.30     &    0.8854     &    0.0090      &     29.59      &     0.9359      &      0.0360 &  30.42  &  0.9175   &  0.0530    \\
\textsc{FTrojan}                 &    41.11    &    0.9885    &    0.0007     &     41.16    &   0.9946      &     0.0006     &     42.28      &     0.9931      &     0.0003  &  42.25  &  0.9904   &  0.0002    \\
\textsc{FIBA}                    &    29.74    &   0.9589     &    0.0083     &    29.69     &    0.9858     &     0.0024     &      29.39     &    0.9755       &     0.0080  &  29.25  &  0.9592   &  0.0057    \\
\textsc{LFBA}                    &    \textbf{43.71}    &   \textbf{0.9943}     &    \textbf{0.0003}     &   \textbf{44.31}      &    \textbf{0.9971}     &     \textbf{0.0001}     &     \textbf{43.54}      &     \textbf{0.9942}      &     \textbf{0.0002} &  \textbf{46.27}  &  \textbf{0.9953}   &  \textbf{0.0001}    \\ \bottomrule
\end{tabular}}
\vspace{-0.2in}
\end{table}

\subsection{Attack Against Defensive Measures}
We evaluate attack robustness of LFBA against the mainstream defenses including Neural Cleanse \cite{nc}, STRIP \cite{strip}, Fine-pruning \cite{fine_pruning} and network inspection.
We also show imperceptible frequency artifacts of LFBA against frequency artifacts inspection \cite{rethink}.
Moreover, we evaluate our attack under preprocessing-based defenses as in \cite{color,ft} to comprehensively illustrate the practical robustness. 

\begin{figure*}
\vspace{-0.2in}
    \centering 
     \begin{subfigure}[b]{0.24\textwidth}
         \centering
\includegraphics[width=\textwidth]{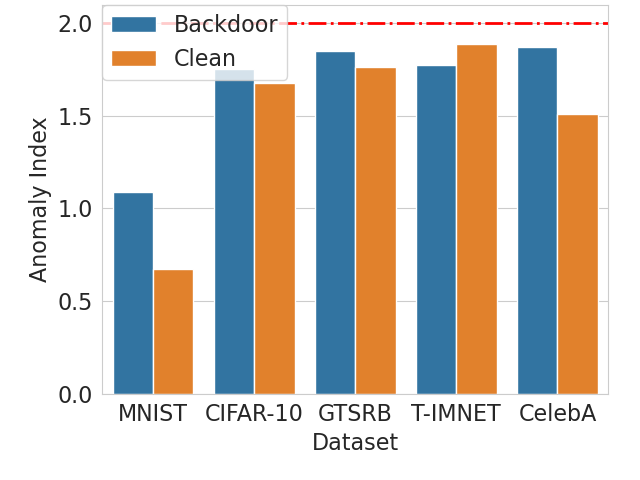}
         \caption{NC}
     \end{subfigure}
     \centering 
     \begin{subfigure}[b]{0.24\textwidth}
         \centering
\includegraphics[width=\textwidth]{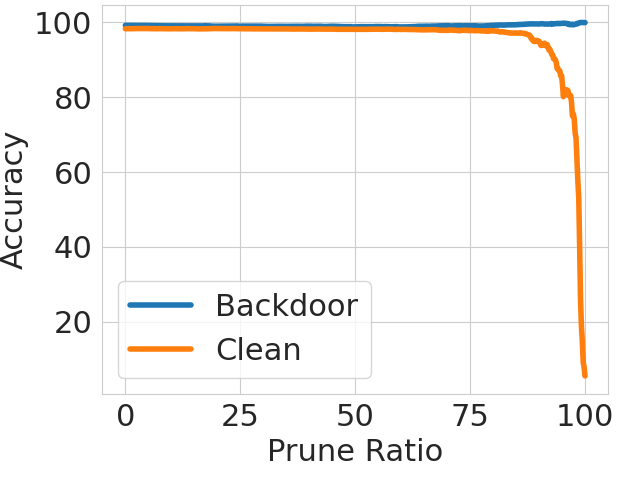}
         \caption{GTSRB}
     \end{subfigure}
     \begin{subfigure}[b]{0.24\textwidth}
         \centering
         \includegraphics[width=\textwidth]{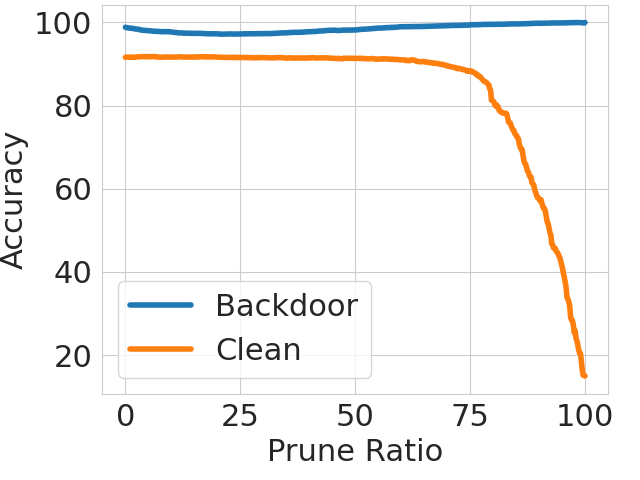}
         \caption{CIFAR-10}
    \end{subfigure}
    \begin{subfigure}[b]{0.24\textwidth}
     \centering
     \includegraphics[width=\textwidth]{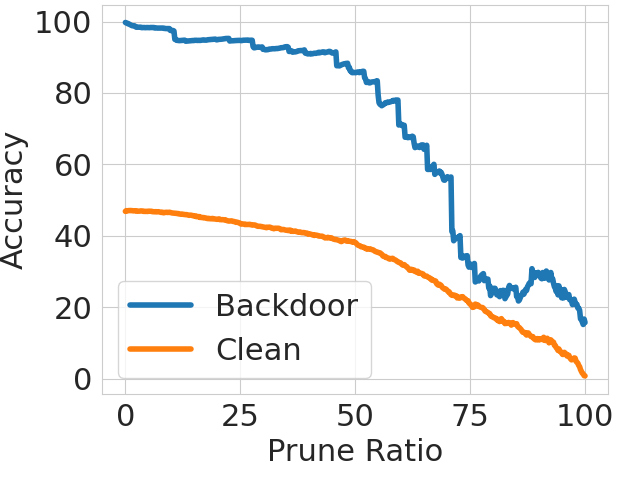}
     \caption{T-IMNET}
    \end{subfigure}
     \caption{(a): The results of LFBA under NC on different datasets; (b)-(d): The attack effectiveness of LFBA against Fine-pruning.}
     \label{nc_fine_pruning}
     \vspace{-0.2in}
\end{figure*}

\noindent\textbf{Neural Cleanse (NC).}
The key intuition of NC is that a backdoor trigger can cause any input misclassified to target label. 
It reverses engineering possible triggers and detects backdoors in the victim model using anomaly index.  
An anomaly index exceeding 2 signifies that the model has been compromised. 
LFBA remains below the threshold and successfully evades the defense across all datasets (see \Cref{nc_fine_pruning} (a)).
We recall that NC focuses on small and fixed patches but LFBA designs trigger in frequency space, wherein it inserts an imperceptible frequency perturbation only causing a minimal change in the entire pixel domain. 
Consequently, trigger spans the entire pixel space, providing considerable natural similarity.

\begin{figure*}
 \vspace{-0.2in}
     \centering 
     \begin{subfigure}[b]{0.19\textwidth}
         \centering
\includegraphics[width=\textwidth]{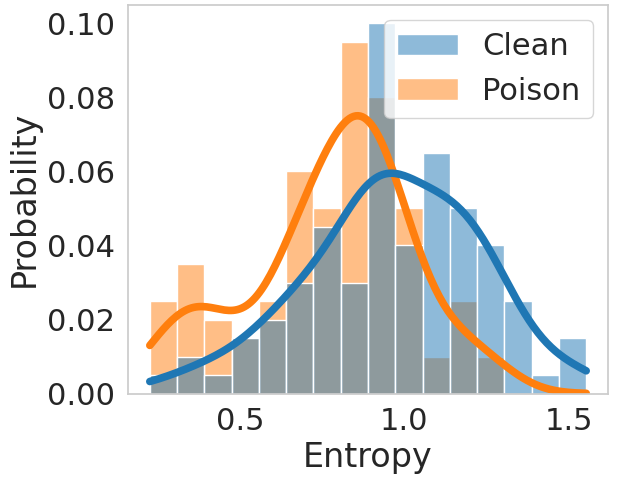}
         \caption{MNIST}
     \end{subfigure}
     \begin{subfigure}[b]{0.19\textwidth}
         \centering
         \includegraphics[width=\textwidth]{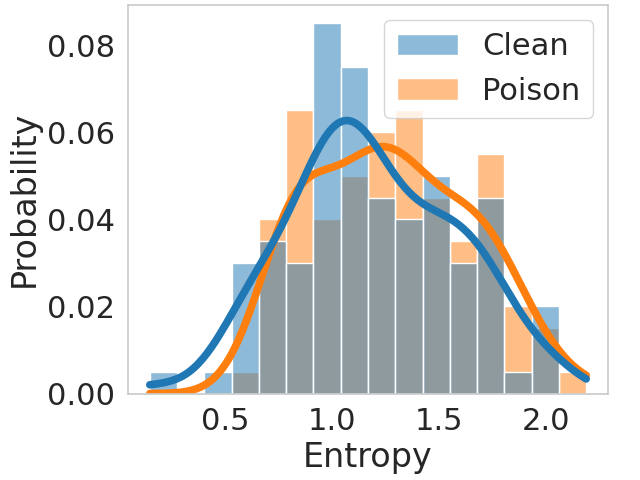}
         \caption{GTSRB}
     \end{subfigure}
     \begin{subfigure}[b]{0.19\textwidth}
         \centering
         \includegraphics[width=\textwidth]{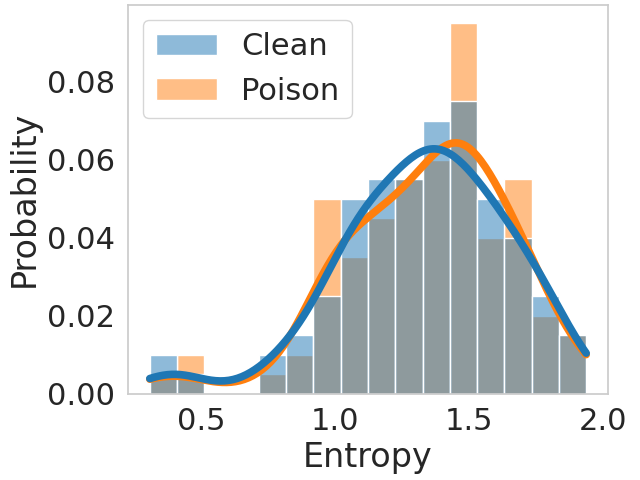}
         \caption{CIFAR-10}
     \end{subfigure}
     \begin{subfigure}[b]{0.19\textwidth}
         \centering
         \includegraphics[width=\textwidth]{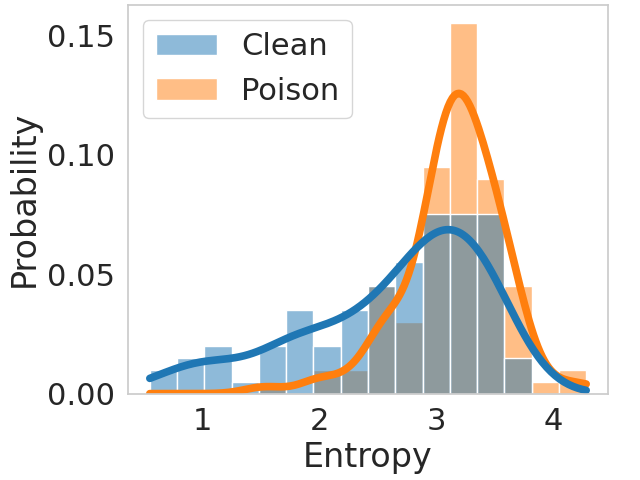}
         \caption{T-IMNET}
     \end{subfigure}
     \begin{subfigure}[b]{0.19\textwidth}
         \centering
         \includegraphics[width=\textwidth]{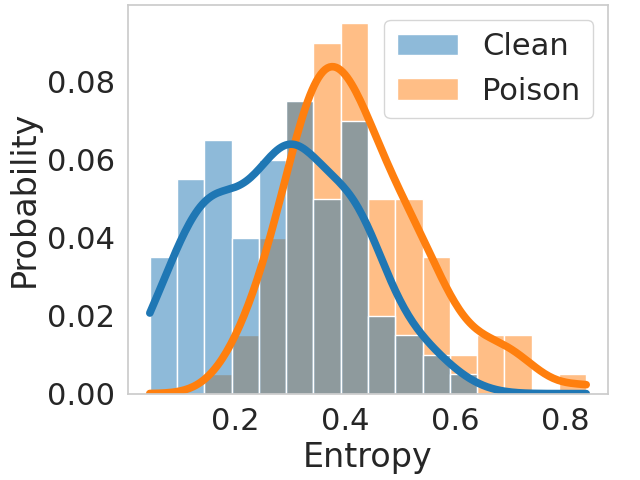}
         \caption{CelebA}
     \end{subfigure}
     \caption{The entropy distributions of LFBA against STRIP under 5 datasets.}
     \label{strip}
     \vspace{-0.2in}
\end{figure*}

\noindent\textbf{STRIP.} It assumes that the predictions given by a backdoored model on poisoned samples consistently tend to be target label and are not easily changed. 
It detects poisoned samples by analyzing the classification entropy after superimposing some random clean samples on the test samples.
We can observe that LFBA achieves almost the same entropy probability distributions as clean samples (see \Cref{strip}), allowing it to circumvent the defense.  
The overlap area of distributions refers to the difficulty of poisoned sample detection. 
For example, the distributions of clean and poisoned samples on CIFAR-10 are almost indistinguishable, indicating that it is hard for STRIP to detect our poisoned samples.
This is so because superimposing random images in spatial space may introduce frequency disparities in poisoned samples. 
Therefore, the predictions of superimposed images will also undergo significant changes, resembling the clean case. 

\noindent\textbf{Fine-pruning.} 
It mitigates backdoor effectiveness by pruning dominant neurons with very low activations via a small clean dataset. 
We test LFBA against Fine-Pruning and demonstrate ACC and ASR with respect to the ratio of pruned neuron on GTSRB, CIFAR-10 and T-IMNET (see \Cref{nc_fine_pruning} (b)-(d)). 
Across all datasets, we see that the ASR is always higher than ACC without any degradation, making backdoor mitigation impossible. 
This suggests that Fine-pruning is ineffective against LFBA.

\begin{figure}[t]
  \centering\scalebox{0.75}{
\includegraphics[width=1\textwidth]{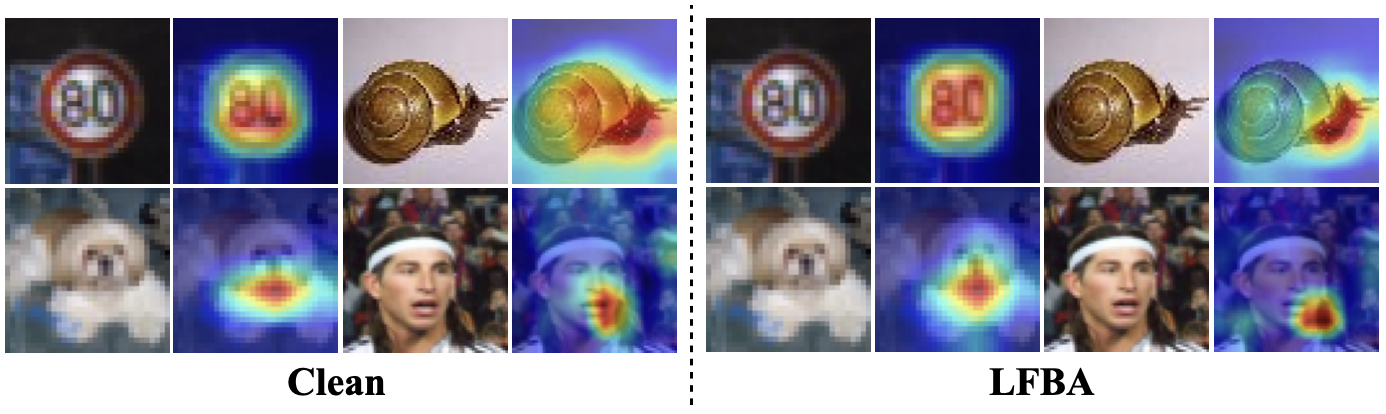}}
  \caption{Visualization of network attention by Grad-CAM on GTSRB, CIFAR-10, T-IMNET and CelebA. Compared to the visualization heatmaps of clean images, LFBA does not introduce any unusual regions.}
  \label{gradcam}
  \vspace{-0.2in}
\end{figure}

\noindent\textbf{Network Inspection.}
We further investigate the impact of LFBA on the attention of the classifier by Grad-CAM \cite{gradcam} and explain why LFBA provides attack robustness under existing defenses. 
Grad-CAM finds the critical regions of input images that mostly activate model’s prediction. 
We showcase visual heatmaps of images on GTSRB, CIFAR-10, T-IMNET and CelebA (see \Cref{gradcam}). 
We observe that LFBA does not introduce anomaly attention area of networks when compared to correct regions across all datasets. 
This is because our trigger is inserted in low-frequency components, which contain the semantics of images, making network attention unaltered.
The results suggest the attack robustness of LFBA against backdoor defenses from a network inspection perspective.

\noindent\textbf{Image Preprocessing-based Defenses.}
We select image preprocessing methods in \cite{ft,color}, including Gaussian filter, Wiener filter, BM3D \cite{bm3d} and JPEG compression \cite{jpeg}, which directly denoise or compress input images. 
We further apply these operations to poisoned test images with various hyperparameters before inference.
The results are shown in \Cref{preprocess_defense}. 
They demonstrate that all transformations are effective to remove trigger effectiveness in FTrojan, which handcrafts mid and high frequency components with anomaly perturbations.
However, LFBA can circumvent these defenses since denoisng transformations and lossy compression do not typically operate on the lower frequency components \cite{lfap}.
We also showcase poisoned images and their frequency disparities (compared to clean images) under preprocessing-based operations in \Cref{fig:visualize_preprocessing} (see \Cref{Exp_of_Robust}) and explain why LFBA is more robust than existing attacks against such image transformations.

\begin{table}[]
\vspace{-0.2in}
\centering
\caption{Attack robustness of various triggers against preprocessing-based defenses. To illustrate the robustness of our low-frequency trigger, we introduce a full-spectrum variant of LFBA for comparison, named LFBA-Full, which searches the trigger across the entire spectrum with same attack settings.}
\label{preprocess_defense} \scalebox{0.9}{
\begin{tabular}{@{}ccccccccc@{}}
\toprule
Attacks $\rightarrow$  & \multicolumn{2}{c}{\textsc{BadNets}} & \multicolumn{2}{c}{\textsc{FTrojan}} & \multicolumn{2}{c}{\textsc{LFBA-Low}} & \multicolumn{2}{c}{\textsc{LFBA-Full}} \\ \cmidrule(l){2-3} \cmidrule(l){4-5} \cmidrule(l){6-7} \cmidrule(l){8-9}
Methods $\downarrow$   & ACC          & ASR          & ACC          & ASR          & ACC            & ASR           & ACC           & ASR          \\ \midrule
Original &    92.02          &    98.78          &     92.53         &        99.82      &       92.91         &       99.88        &    91.21           &  99.73            \\
Gaussian Filter ($w=(3,3)$)     &             66.17          &    15.11          &     67.80         &        6.47      &       72.23         &       98.52        &    71.46           &  7.81           \\
Gaussian Filter ($w=(5,5)$)      &       39.81          &    6.88          &     45.03         &        3.25      &       53.54         &       97.27        &    49.07           &  3.59             \\
Wiener Filter ($w=(3,3)$)      &     69.53          &      96.02        &     69.11         &        10.54      &       71.92         &       98.22        &    71.61           &  6.93            \\
Wiener Filter ($w=(5,5)$)       &      52.18          &    90.81          &     49.20         &        5.28      &       52.03         &       95.65        &    50.90           &  3.66           \\
BM3D ($\sigma=0.5$)    &         87.39          &    98.44          &     87.34         &        15.84      &       88.31         &       99.09        &    87.08           &  13.58    \\
BM3D ($\sigma=1.0$)    &        86.03          &    94.07          &     86.40         &        19.33      &       86.70         &       98.04        &    86.64           &  18.69    \\
JPEG (quality = $90\%$)    &          88.98          &    97.85          &     89.22         &        9.36      &      82.72          &       89.75        &    89.18           &  9.54        \\
JPEG (quality = $50\%$)    &        78.84          &    92.59          &     79.66         &        8.58      &       76.19         &       75.93        &    76.31           &  8.54        \\ \midrule
Average ASR    &                 &    73.97          &            &        9.83      &               &       \textbf{94.06}        &              &  9.04 \\
\bottomrule
\end{tabular}}
\vspace{-0.2in}
\end{table}

\begin{figure}[t]
\vspace{0.1in}
  \centering\scalebox{0.8}{
  \includegraphics[width=1\textwidth]{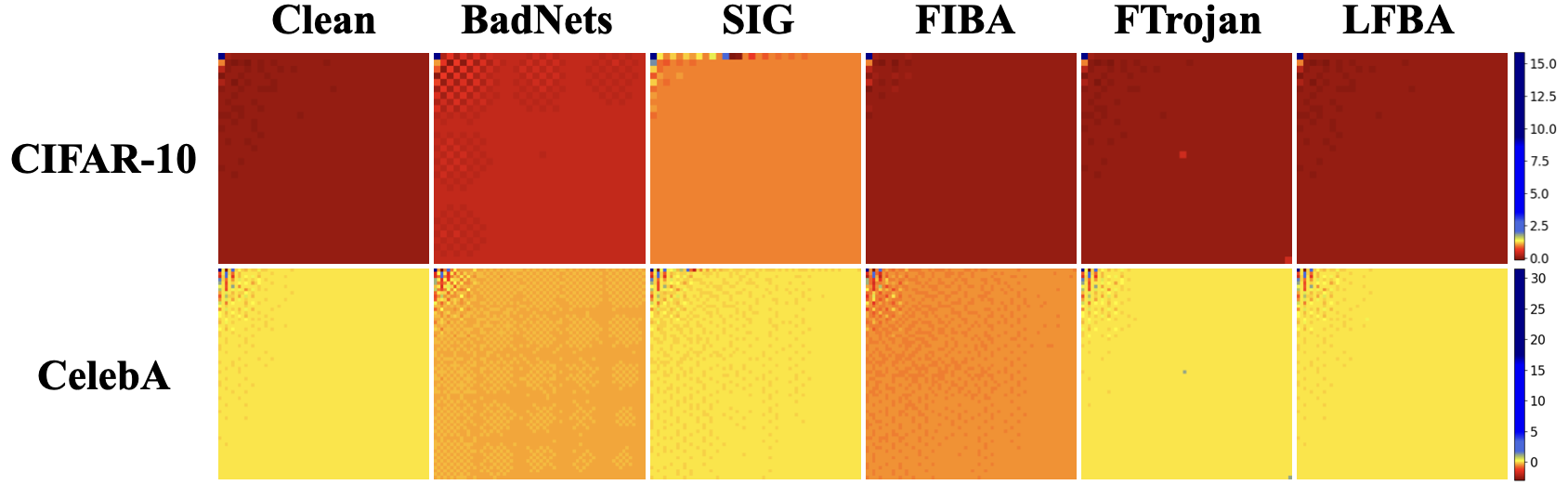}}
  \caption{Visualization of DCT spectra between clean and posioned samples under various spatial and frequency attacks on two different input-space datasets including CIFAR-10 ($32\times32$) and CelebA ($64\times64$). We randomly select 10000 samples from each dataset and showcase the averaged spectrum results. }
  \label{freq_inspection}
  \vspace{-0.1in}
\end{figure}

\noindent\textbf{Frequency Artifacts Inspection.}
We consider the same frequency artifacts inspection method as in \cite{rethink}.
In \Cref{freq_inspection}, we compare the frequency spectra between clean and poisoned images with various spatial and frequency triggers. 
We can see that current spatial backdoors, such as SIG, have larger disparities than frequency backdoors although they could achieve perceptual similarity.
Moreover, state-of-the-art frequency backdoors also introduce anomaly frequency artifacts.
It is worth noting that two spikes lie in central and bottom-right regions in FTrojan's spectra and a noticeable color shift exists in FIBA's.
However, LFBA spectra closely resemble clean images on both datasets (exhibiting similarly smooth spectra as clean samples).
According to \cite{burton1987color,tolhurst1992amplitude,rethink}, our poisoned samples exhibit the same frequency properties as natural images due to dual-space stealthiness. 
Therefore, frequency inspection is ineffective to detect anomaly artifacts of LFBA.
 
In conclusion, a wide range of experimental results empirically demonstrate that \emph{LFBA can elude or significantly degrade the performance of the state-of-the-art defenses in both spatial and frequency spaces} even when both the model and defense strategy are unknown.
Besides, our frequency trigger is \emph{resilient to image preprocessing-based defenses, which provides more robustness than existing attacks.} 
The results also indicate the pivotal role of LFBA in bolstering the security of machine learning systems. 

\subsection{Ablation Study}
We here analyze several hyperparameters that are critical for the LFBA performance including frequency stealthiness budget $\epsilon$, number of manipulated bands $n$ and poison ratio $\rho$. 

\begin{figure}[t]
  \centering\scalebox{0.95}{
  \includegraphics[width=1\textwidth]{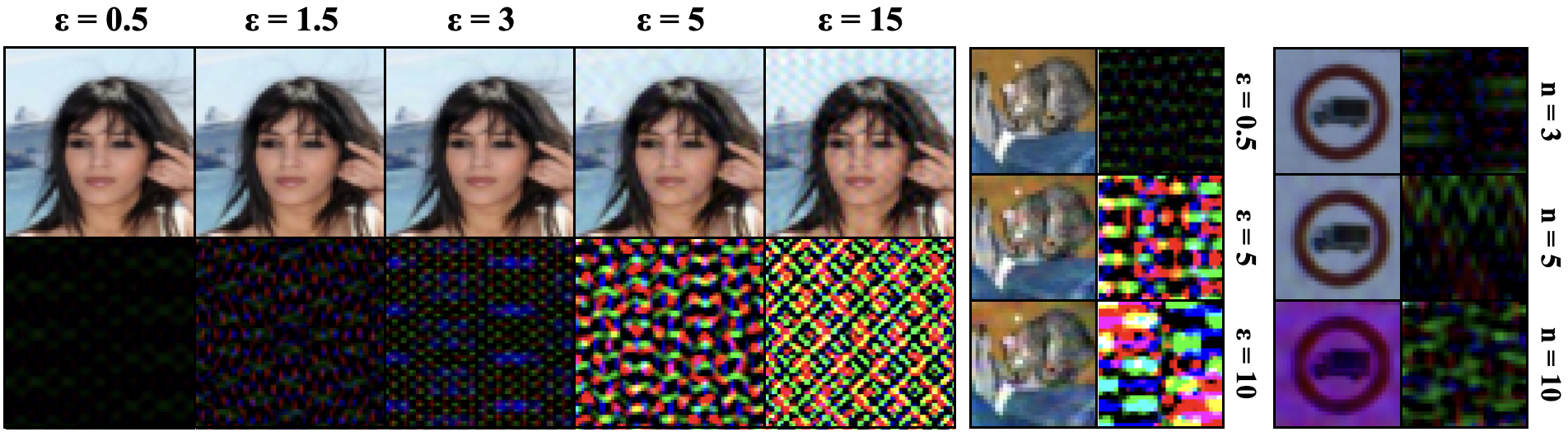}}
  \caption{Visualization of LFBA poisoned images and triggers under different $\epsilon$ and $n$. The pixel value of triggers is amplified by 30$\times$.}
  \label{visual_e_n}
  \vspace{-0.2in}
\end{figure}

\begin{figure*}
\vspace{-0.2in}
     \centering 
     \begin{subfigure}[b]{0.35\textwidth}
         \centering
         \includegraphics[width=\textwidth]{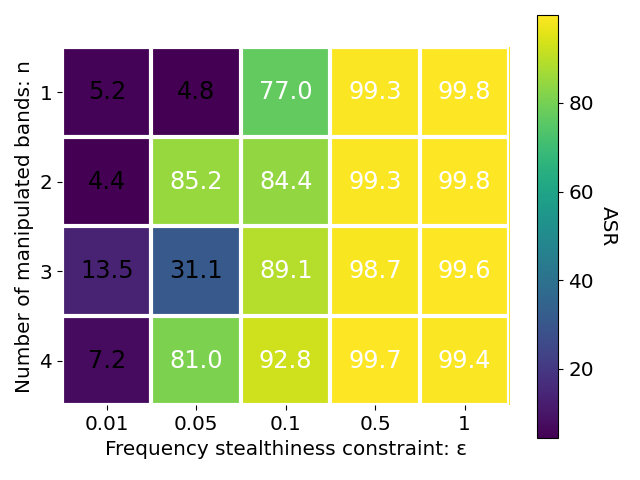}
         \caption{GTSRB}
     \end{subfigure}
     \begin{subfigure}[b]{0.35\textwidth}
         \centering
         \includegraphics[width=\textwidth]{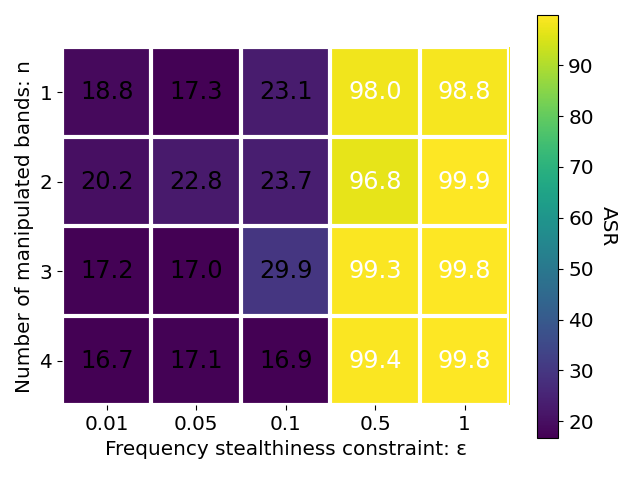}
         \caption{CIFAR-10}
     \end{subfigure}
    
     \caption{The impact of $\epsilon$ and $n$.}
     \label{epsilon_n}
     \vspace{-0.2in}
\end{figure*}

\noindent\textbf{Frequency stealthiness constraint $\epsilon$ and number of manipulated band $n$.}
$\epsilon$ restrains the maximum perturbation of each frequency band while $n$ controls the number of manipulated frequency bands.  
We visualize the impact of $\epsilon$ and $n$ on the poisoned images (see \Cref{visual_e_n}).
If $\epsilon$ is set too large, the poisoned image may be easily recognized (i.e., lacking stealthiness) upon human inspection in the pixel domain. 
Moreover, this could introduce distinguishable disparities in the frequency space.
On the other hand, setting $\epsilon$ too small results in the trigger having a low proportion of features in dual spaces. 
In this sense, the classifier will encounter difficulty in catching and learning these trigger features, yielding a drop of attack effectiveness.  
\Cref{epsilon_n} illustrates the influences of $\epsilon$ and $n$ on the attack effectiveness among the tasks.  
The ASRs of LFBA decline significantly and eventually fall below 20\% as we continuously decrease $\epsilon$ to $0.01$, in which evidences can be seen in GTSRB under various $n$, while there is a drastic drop occurs from $\epsilon=0.5$ to $\epsilon=0.1$ in CIFAR-10. 
We notice that increasing $n$ enhances the effectiveness of LFBA.
For instance, the ASR increases from $77\%$ to $92.8\%$ under $\epsilon=0.1$ when $n$ increases from $1$ to $4$ in GTSRB.
We also observe that a large $\epsilon$ allows single injection in frequency band to achieve a high ASR.
For instance, selecting $\epsilon=1$ and $n=1$ to perform our attack can achieve nearly $100\%$ ASR. 
However, such an attack setup could compromise frequency stealthiness.
Thus, it's crucial to consider a balance between dual-space stealthiness and attack effectiveness before conducting a LFBA attack.

\noindent\textbf{Poison ratio $\rho$.} We test LFBA's effectiveness across a range of $\rho$ values from 0.1\% to 10\%, and verify that LFBA is robust to this hyperparameter. 
Details can be seen in \Cref{appendix:poison ratio}.

%% file: 6conclusion.tex
\section{Conclusion}
We propose a robust black-box backdoor attack by inserting imperceptible perturbations in the frequency domain.
Our method provides perceptual similarity in the pixel domain and imperceptibility in the frequency domain.
We leverage SA to optimize the trigger to satisfy the attack effectiveness and dual-space stealthiness.
The empirical experiments demonstrate that LFBA can achieve a practical attack robustness to evade SOTA defenses in both spatial and frequency domains as well as image preprocessing-based defenses. 

\noindent\textbf{Discussion and Limitation.}
In this work, we concentrate on various computer vision tasks, which have been the focus of numerous existing works \cite{wanet,dynamic,lira,marksman,ft}. 
In the future, we intend to expand the scope of this work to other SOTA model architectures, e.g., diffusion and transformer models.
The trigger search process is executed in a hybrid GPU-CPU environment during trigger evaluation and optimization phases.
It deserves further efforts to design a GPU-accelerated SA to minimize data transmission across hardware, thus improving the efficiency of our proposed LFBA.
To counter the proposed attacks, we hope to devise more robust frequency defenses that can surpass the current assumption of both spatial and frequency domain-based backdoor attacks. 
We also hope this work can inspire follow-up studies that enhance the security and robustness of machine learning systems against LFBA through a frequency perspective.

%% file: 8appendix.tex
\newpage
\appendix

\section{Appendix}

\subsection{Details of Computer Vision Tasks}
\label{appendix:task}

\begin{table}[]
\centering
\caption{The summary of tasks, and their corresponding models.}
\label{task} \scalebox{0.7}{
\begin{tabular}{@{}cccccc@{}}
\toprule
Task                          & Dataset  & \# of Training/Test Images & \# of Labels & Image Size                                        & Model Architecture               \\ \midrule
Handwritten Digit Recognition & MNIST    & 60,000/10,000                & 10           & 28$\times$28$\times$1   & 3 Conv $+$ 2 Dense \\
Object Classification         & CIFAR-10 & 50,000/10,000                & 10           & 32$\times$32$\times$3   & PreAct-ResNet18 \\
Traffic Sign Recognition      & GTSRB    & 39,209/12,630                & 43           & 32$\times$32$\times$3   & PreAct-ResNet18                           \\
Object Classification         & Tiny-ImageNet & 100,000/10,000                        & 200           & 64$\times$64$\times$3 & ResNet18                        \\ 
Face Attribute Recognition         & CelebA & 162,770/19,962               & 8           & 64$\times$64$\times$3 & ResNet18                        \\ \bottomrule
\end{tabular}}
\vspace{-0.2in}
\end{table}

\begin{figure*}
    \centering
    \begin{subfigure}[b]{0.35\textwidth}
         \centering
        \includegraphics[width=\textwidth]{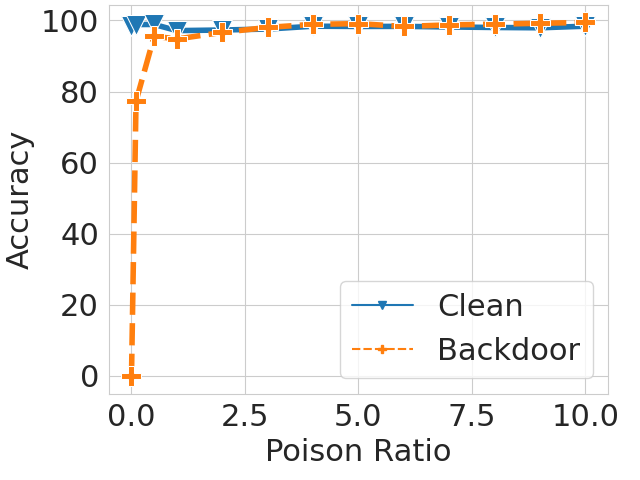}
         \caption{GTSRB}
    \end{subfigure}
    \begin{subfigure}[b]{0.35\textwidth}
         \centering
        \includegraphics[width=\textwidth]{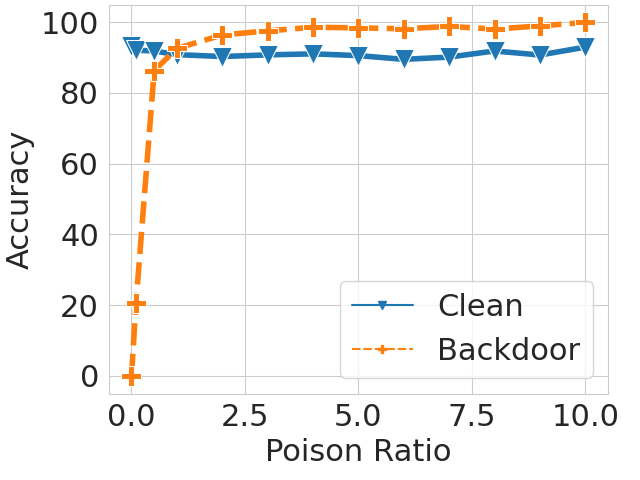}
         \caption{CIFAR-10}
    \end{subfigure}
    \caption{The impact of attack effectiveness under a wide range of poison ratios (\%).}
    \label{fig_poison_ratio}
    \vspace{-0.2in}
\end{figure*}

\subsection{Poison Ratio $\rho$}
\label{appendix:poison ratio}

$\rho$ is the fraction of poisoned samples in the training dataset of the adversary. 
We test the attack effectiveness under different $\rho$ varying from 0.1\% to 10\%.
Although we increase $\rho$ from a wide range, LFBA does not harm the ASR of the victim models. 
As stated in \Cref{fig_poison_ratio}, this fraction setting cannot degrade the ACC and meanwhile, we would like to examine the lower bound of the fraction that LFBA’s effectiveness can withstand.
Even when $\rho$ is $0.1\%$, LFBA can still provide a high ASR, around 80\% for GTSRB.
We also find that sensitivities to poison ratio can vary among tasks.
In CIFAR-10, LFBA achieves above $86\%$ ASR under $\rho=0.5\%$ while it drops rapidly, around $20\%$, when $\rho$ reduces to $0.1\%$.

\begin{figure*}[t]
    \centering
    \begin{subfigure}{1.0\textwidth}
         \centering\scalebox{0.35}{
\includegraphics{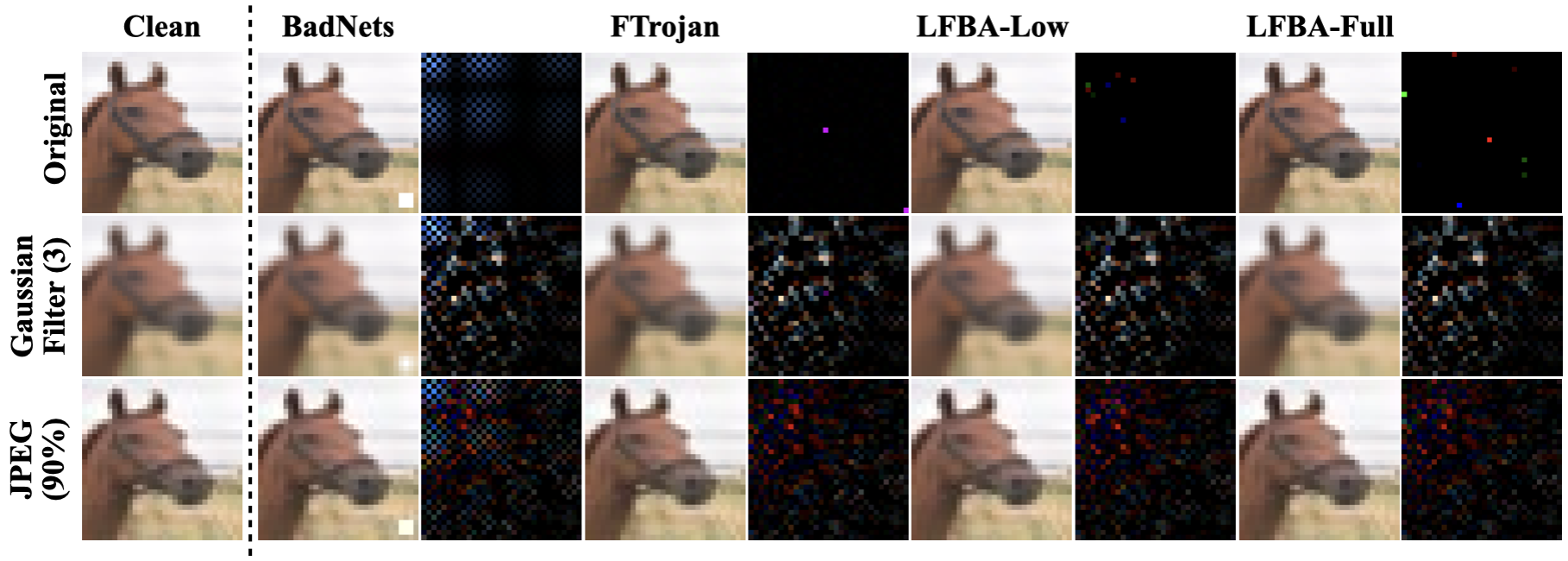}}
        \caption{CIFAR-10}
    \end{subfigure}
    \begin{subfigure}{1.0\textwidth}
         \centering\scalebox{0.35}{
\includegraphics{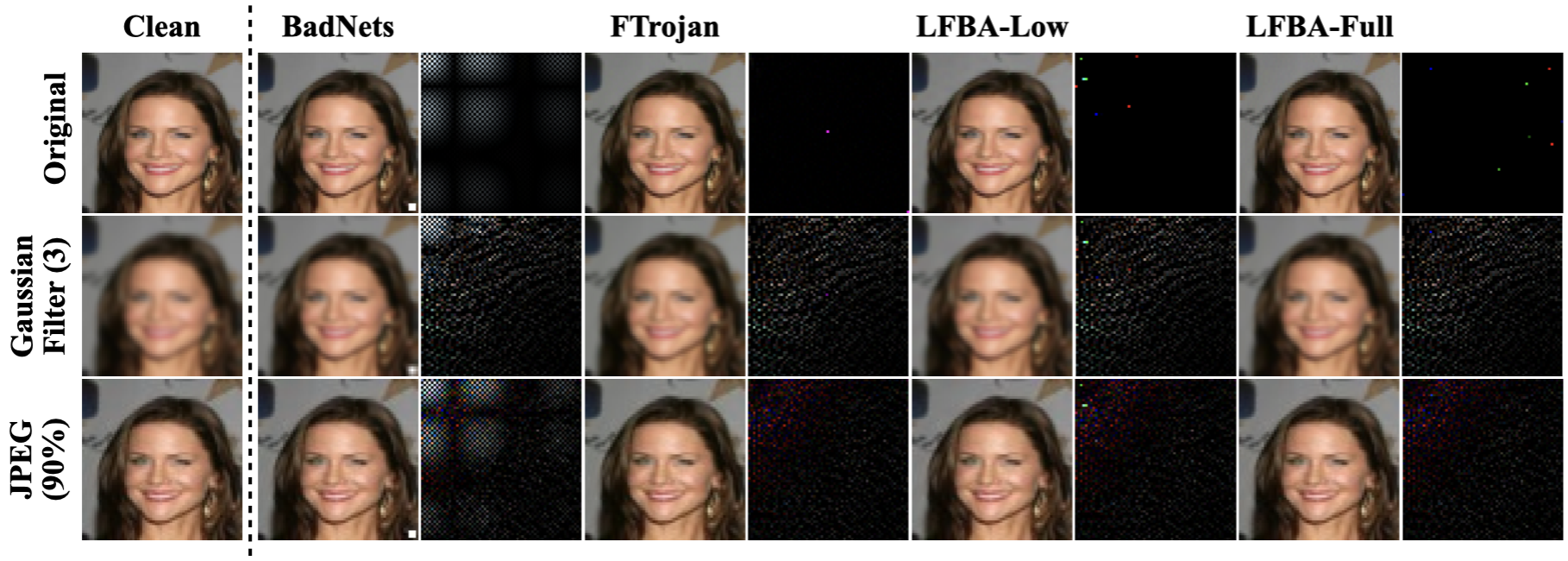}}
         \caption{CelebA}
    \end{subfigure}
    \caption{Comparison of poisoned images with their corresponding frequency disparities (amplified by $5\times$) to clean images of existing attacks under different image preprocessing-based defenses.
    Each frequency disparities spectrum is calculated based on the original clean image's spectrum.
    These image transformations can effectively remove the trigger pattern through frequency domain, while the disparities spectrums of our LFBA-Low attack still contain original backdoor patterns. }
    \label{fig:visualize_preprocessing}
    \vspace{-0.2in}
\end{figure*}

\subsection{Explanation of Robustness through Frequency Perspective.}
\label{Exp_of_Robust}

We showcase poisoned images and their frequency disparities (compared to clean images) under the image transformations in \Cref{fig:visualize_preprocessing}.
We can see that the frequency disparities of BadNets remain similar to the original ones after JPEG compression while the Gaussian filter destroys the BadNets patterns on both datasets. 
This proves the fact, as shown in \Cref{preprocess_defense}, that BadNets is effective against JPEG compression but fails to survive after Gaussian filtering.
For FTrojan and LFBA-Full, we cannot see any frequency patterns after these transformations. 
However, the frequency disparities of 
LFBA-Low can be clearly seen even after such operations, indicating our low-frequency attack is robust against preprocessing-based defenses. 
We note that low-frequency components exhibit greater resilience to image transformations than mid- and high-frequency components.